\documentclass[lettersize,journal]{IEEEtran}
\usepackage{amsmath,amsfonts}
\usepackage{array}
\usepackage[caption=false,font=normalsize,labelfont=sf,textfont=sf]{subfig}
\usepackage{textcomp}
\usepackage{stfloats}
\usepackage{url}
\usepackage{verbatim}
\usepackage{graphicx}
\usepackage{cite}
\usepackage{booktabs}
\usepackage{multirow}
\usepackage{makecell}
\usepackage{threeparttable}
\usepackage[colorlinks=true,
            linkcolor=blue,
            citecolor=blue,
            urlcolor=blue]{hyperref}
\usepackage[ruled,lined]{algorithm2e}
\SetKwComment{Comment}{$\triangleright$\ }{}

\begin{document}

\title{LaCoVL-FER: Landmark-Guided Contrastive Learning Network with Vision-Language Enhancement for Facial Expression Recognition}

\author{Jiaxin Wang, Muwei Jian, Hui Yu,~\IEEEmembership{Senior Member,~IEEE}, Junyu Dong,~\IEEEmembership{Member,~IEEE}, and Yifan Xia
\thanks{This work was supported in part by the National Natural Science Foundation of China under Grant 62501362, in part by Shandong Provincial Natural Science Foundation under Grant ZR2024QF018, and in part by Taishan Scholars Program of Shandong Province (tstp20250536). \emph{Corresponding author: Yifan Xia.}}
\thanks{J. Wang and Y. Xia are with the School of Airspace Science and Engineering, Shandong University, Weihai, China (e-mail: xiayifan@sdu.edu.cn).}
\thanks{M. Jian is with the School of Computer Science and Technology, Shandong University of Finance and Economics, Jinan, China (e-mail: jianmuweihk@163.com).}
\thanks{H. Yu is with the School of Psychology and Neuroscience, University of Glasgow, Glasgow, U.K. (e-mail: hui.yu@glasgow.ac.uk).}
\thanks{J. Dong is with the Faculty of Information Science and Engineering, Ocean University of China, Qingdao, China (e-mail: dongjunyu@ouc.edu.cn).}}


\maketitle
\begin{abstract}
Facial Expression Recognition (FER) in the wild requires models to identify subtle expression cues under large variations in pose, occlusion, illumination, and identity. Recent FER methods improve robustness by introducing visual attention, facial landmarks, or vision-language models as auxiliary priors. However, these priors are typically integrated in a static manner, failing to capture instance-specific facial variations, thereby resulting in severe attention redundancy and representation instability. To address this issue, we propose LaCoVL-FER, a landmark-guided contrastive learning network with vision-language enhancement for FER, which shifts FER from static prior injection to sample-adaptive prior refinement. Specifically, a Landmark-Guided Adaptive Encoder (LGAE) calibrates regional appearance features with landmark geometry through Bi-branch Gated Cross Attention (BGCA), suppressing noisy responses and producing expression-relevant representations. In parallel, a Vision-Language Enhancement Strategy (VLES) refines the generalizable visual features from a frozen CLIP image encoder into expression-specific visual representations. Based on them, an Expression-Conditioned Prompting (ECP) mechanism adapts fixed class-level textual prompts from the frozen CLIP text encoder into instance-aware textual representations. The resulting visual-textual representations are aligned as adaptive semantic priors to enhance robustness and generalization. Quantitative and qualitative experiments show that LaCoVL-FER outperforms state-of-the-art methods on RAF-DB, FERPlus, and AffectNet. The code is available at https://github.com/ylin06804/LaCoVL-FER.
\end{abstract}


\section{Introduction}
\IEEEPARstart{F}{acial} expressions are a fundamental way for humans to convey emotions~\cite{ref1} and play an irreplaceable role in interpersonal communication~\cite{ref2}. Facial Expression Recognition (FER) aims to classify facial images into predefined emotion categories. It has been widely applied in psychological studies~\cite{ref3}, driver fatigue detection~\cite{ref7}, and healthcare services~\cite{ref5}. It also promotes the development of affective computing~\cite{ref6} and human-computer interaction~\cite{ref4}. Although deep models have improved FER, recognizing expressions in the wild remains difficult. Real images often contain occlusion, pose changes, illumination variation, background noise, and identity differences. These factors make expression cues weak, local, and unstable. Therefore, the key problem is not only how to extract stronger visual features, but also how to decide which facial regions and which category cues are reliable for each input image.

Existing FER methods mainly address this problem from three directions. Visual attention methods learn to highlight expression-related regions from appearance features~\cite{ref8,ref9}. Landmark-based methods introduce facial geometry to guide the model toward important facial components, such as eyes, eyebrows, nose, and mouth~\cite{ref15,ref16,ref17}. Vision-language methods use class-level textual prompts from pretrained vision-language models, such as CLIP~\cite{ref29}, to provide category-level guidance~\cite{ref24,ref25}. These methods have greatly improved FER in real-world scenarios. They show that appearance features, facial geometry, and class-level textual cues are all valuable priors for expression recognition.

\begin{figure}[!t]
\centering
\includegraphics[width=3.5in,trim=18 18 18 15,clip]{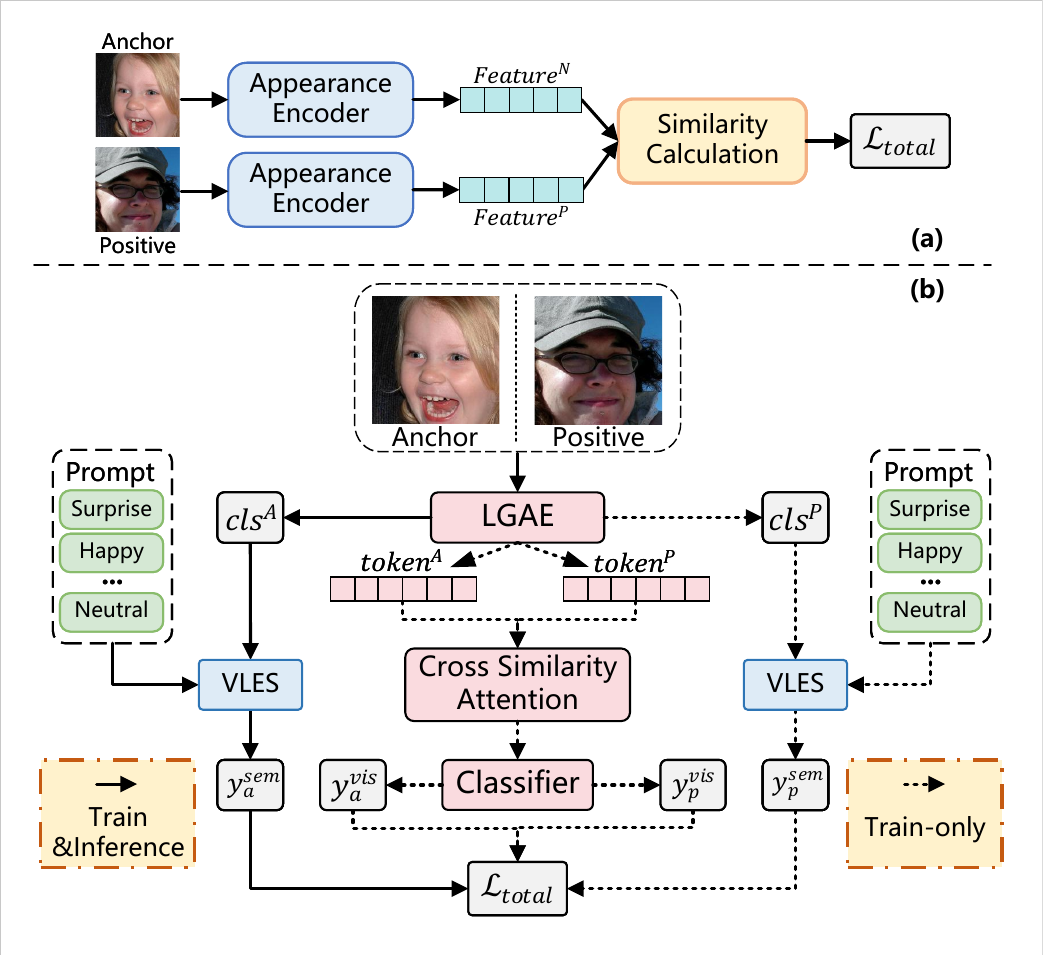}
\caption{Comparison between conventional contrastive learning framework and the proposed LaCoVL-FER. (a) The conventional contrastive learning framework learns visual feature based solely on appearance features and similarity objectives, which may suffer from attention redundancy and unstable representations. (b) The proposed LaCoVL-FER integrates geometric priors from facial landmarks and semantic priors from a vision-language model to learn more discriminative and robust representations for facial expression recognition.}
\label{fig_1}
\end{figure}

However, most existing methods still utilize these cues in a direct or static way. Appearance attention is often learned only from visual responses, making it vulnerable to background noise, hair, or local occlusion~\cite{ref10,ref11,ref12}. Landmark-based methods usually inject geometric features as additional inputs, but directly employing all features without adaptive selection may overweight irrelevant regions since not all landmarks contribute equally to every expression~\cite{ref18}. Vision-language methods often rely on fixed templates or generic visual-textual features, which treat vision and language as independent modalities and lack deep, sample-adaptive cross-modal interactions, failing to match the subtle expression shown by the current face~\cite{ref30,ref31,ref32}. Consequently, these methods do not fully answer a more basic question: how should each prior be adaptively used and tailored for the current sample?

Based on our observations, despite the remarkable progress made so far, FER in real-world scenarios remains challenging for two main reasons. On the one hand, different expressions often exhibit subtle and localized variations (e.g., between fear and surprise, or disgust and anger), leading to high inter-class similarity and thus distracting attention from task-relevant regions. On the other hand, the same expression may exhibit significant intra-class variations due to differences in identity, pose, illumination, and occlusion, resulting in inconsistent and unstable focus on critical regions. If these prior cues are directly or rigidly trusted without sample-adaptive adaptation, the model will inevitably suffer from severe attention redundancy from irrelevant regions and representation instability from intra-class variations. Therefore, we argue that robust in-the-wild FER requires dynamic, sample-conditioned representation refinement. Instead of static injection, the model should enact sample-adaptive integration to transform raw appearance, geometric, and textual cues into optimized, instance-specific representations.

To address these issues, we propose a novel landmark-guided contrastive learning network with vision-language enhancement for FER, termed LaCoVL-FER. As illustrated in Fig.~\ref{fig_1}(a), the conventional contrastive learning framework~\cite{ref23,ref73} typically extracts generic features using a visual appearance encoder and optimizes them with similarity-based objectives, which fails to handle sample-specific variations. In contrast, the proposed LaCoVL-FER shown in Fig.~\ref{fig_1}(b) implements the sample-adaptive representation refinement paradigm by jointly exploiting geometric priors from facial landmarks and semantic priors from a vision-language model. Specifically, for regional feature optimization, we design a Landmark-Guided Adaptive Encoder (LGAE) with a Bi-branch Gated Cross Attention (BGCA) mechanism to adaptively calibrate the interactions between landmark-based geometry and visual appearance features, thereby producing expression-relevant visual representations while filtering out irrelevant visual redundancy. On top of this, for category-level feature optimization, we further develop a Vision-Language Enhancement Strategy (VLES) integrated with an Expression-Conditioned Prompting (ECP) mechanism. The VLES leverages the refined regional features to dynamically modulate the CLIP-derived generalizable visual features, generating expression-specific visual representations. Based on these optimized visual cues, the ECP mechanism conditionally adapts the textual features of fixed class-level prompts. In this way, the textual semantics can adapt dynamically to each input sample, yielding instance-aware textual representations as tailored semantic priors to further enhance the robustness and generalization of FER. The major contributions of this work are summarized as follows:
\begin{enumerate}
\item[1)] We propose a novel LaCoVL-FER framework for in-the-wild FER. Unlike existing methods that rely on static prior injection, our model dynamically integrates geometric and semantic priors to implement sample-adaptive prior refinement, effectively eliminating the redundant attention and representation instability caused by uncalibrated prior.
\item[2)] We design a Landmark-Guided Adaptive Encoder (LGAE) with a Bi-branch Gated Cross Attention (BGCA) mechanism to introduce geometric priors to focus on key facial regions and suppress noise interference.  Furthermore, we present a Vision-Language Enhancement Strategy (VLES) with an Expression-Conditioned Prompting (ECP) mechanism to generate expression-specific visual and instance-aware textual representations, aligning these CLIP-derived representations as semantic priors to enhance the robustness and generalization of FER. 
\item[3)] Extensive experimental results on three representative real-world FER datasets demonstrate the effectiveness of the proposed method, which achieves state-of-the-art performance.
\end{enumerate}

The rest of this paper is organized as follows. Section II reviews the related work on FER. In Section III, we describe the proposed method. Experimental results on public datasets are presented in Section IV. Finally, we conclude the paper and discuss future work in Section V.

\section{Related Work}
\subsection{Facial Expression Recognition}
Facial Expression Recognition (FER) has achieved remarkable progress with the development of deep learning. Early FER methods mainly relied on handcrafted descriptors, such as Local Binary Patterns (LBP)~\cite{ref33} and Gabor kernels~\cite{ref34}. Although these methods can capture certain local facial patterns, their representation capability is limited and they are difficult to generalize to complex real-world scenarios.

With the rise of deep neural networks, CNN-based FER methods~\cite{ref35,ref36,ref37,ref38,ref39,ref40} have substantially improved facial representation learning through various strategies for robust visual modeling. For example, EfficientFace~\cite{ref39} integrates a deep convolutional feature extractor with channel-spatial modulators to capture both global and local facial representations. MRAN~\cite{ref40} extracts affective information by modeling multi-level relationships between global and local features. Although these methods improve model robustness to some extent, CNNs are still inherently constrained by local convolutional operations, which limits their ability to model global structural relationships and long-range dependencies.

To address this issue, recent studies have introduced Vision Transformer (ViT)~\cite{ref43} to model long-range dependencies and highlight key facial regions. For example, RAN~\cite{ref8} and DSAN~\cite{ref9} improve FER by introducing attention mechanisms to suppress irrelevant responses and emphasize critical local facial expression cues. Building on this line, VTFF~\cite{ref10} generates two feature maps through a dual-branch CNN and models their relationships via self-attention. TransFER~\cite{ref11} further combines multi-branch local CNNs with ViT to extract attention information, and enhances feature diversity through a multi-attention dropout module. APViT~\cite{ref58} further strengthens key information while reducing redundant responses and computational cost through attention pooling. In addition, DCS/QCS~\cite{ref12} introduce contrastive learning and cross similarity attention to enhance feature diversity and robustness. Despite these advances, most of these methods still primarily rely on visual appearance cues, making them prone to assigning excessive attention to task-irrelevant regions in complex in-the-wild scenarios with occlusion, pose variation, and noise interference.

To alleviate these limitations, some studies introduce geometric priors into FER. For instance, POSTER~\cite{ref15}, POSTER++~\cite{ref16}, and LA-Net~\cite{ref17} exploit geometric cues provided by MobileFaceNet~\cite{ref45} to guide attention toward key facial regions, thereby improving robustness to some extent. However, not all facial landmarks contribute equally to FER. Directly aggregating all landmark-related features without an adaptive selection mechanism may introduce redundant or weakly relevant geometric cues, thereby over-emphasizing irrelevant regions, under-emphasizing discriminative ones. This observation motivates us to design a Landmark-Guided Adaptive Encoder (LGAE) equipped with the proposed Bi-branch Gated Cross Attention (BGCA) mechanism for adaptive visual-geometric feature fusion.

\subsection{Vision-Language Pre-training Models in FER}
Vision-Language Pre-training (VLP) models align image and text representations into a shared semantic space, thereby significantly improving zero-shot generalization and promoting the development of multimodal learning. Among them, CLIP~\cite{ref29} is one of the most representative frameworks.

Recently, researchers have begun to incorporate VLP models into FER to exploit semantic priors, thereby alleviating attention instability and enhancing expression understanding. However, owing to the subtle and fine-grained nature of facial expressions, directly transferring generic VLP models is often insufficient for FER. To improve semantic modeling in this context, CLIPER~\cite{ref24} adopts multiple expression-related textual descriptions as semantic supervision and employs a two-stage training strategy to learn more fine-grained and interpretable expression representations. CEPrompt~\cite{ref25} further injects affective concept prompts into facial appearance representations, capturing subtle expression variations through an emotion conception-guided visual adapter, knowledge distillation, and a conception-appearance tuner. Beyond prompt design, VLCA~\cite{ref30} combines a learnable visual encoder with semantic priors provided by frozen CLIP visual and textual encoders, and enhances expression representations through cross-modal interaction. TPRD~\cite{ref31} further semantically aligns regional textual prompts with global facial features and adaptively fuses the contributions of different facial regions to the global representation. More recently, MMPL-FER~\cite{ref32} addresses the modality gap between image and text features, as well as the limitations of fixed textual prompts in CLIP-based FER, by integrating emoji image prompts and semantic textual prompts generated by large language models to enhance fine-grained semantic alignment. Along a similar direction, ICoCO~\cite{ref69} further introduces emotion concept prompts distilled from large language models to establish concept-appearance coupling between visual and textual representations, achieving multi-granularity visual-semantic alignment.

Despite achieving encouraging progress, they still suffer from several limitations. First, they usually directly employ the generic visual features from VLP models, making it difficult to learn expression-specific visual representations. Moreover, most of them rely on fixed templates, predefined prompts, or limited prompt adaptation, which restricts the expressiveness and contextual sensitivity of the language modality. To address these limitations, we propose a Vision-Language Enhancement Strategy (VLES), which improves FER by refining CLIP-derived visual features into expression-specific representations and utilizing an Expression-Conditioned Prompting (ECP) mechanism to adapt textual features into more instance-aware representations.

\section{The Proposed Method}
\subsection{Preliminaries: Cross Similarity Learning}
To exploit category-consistent expression information, our framework adopts the Cross Similarity Learning (CSL) module as established by DCS/QCS~\cite{ref12}. As illustrated in Fig. 2, CSL functions as a baseline contrastive mechanism that models token-level semantic correspondences between paired images ($x_a$ and $x_p$) belonging to the same expression class. By capturing these cross-sample feature dependencies, the module suppresses identity-related variations and background noise to provide robust baseline representations. Concretely, the local token sequences $X_a$ and $X_p$ generated by our Landmark-Guided Adaptive Encoder are mapped into a shared embedding space. Following~\cite{ref12}, these features are mutually aligned via cross similarity attention blocks to guide the token-level relationship modeling. Subsequently, the interaction-enhanced features are concatenated and aggregated through a standard ViT encoder to yield the corresponding baseline visual predictions $y_a^{vis}$ and $y_p^{vis}$, serving as the technical baseline upon which our primary methodological extensions are constructed.

\subsection{Overview}
The overall architecture of the proposed LaCoVL-FER is illustrated in Fig. 2. Given paired input face images consisting of an anchor $x_a$ and a positive sample $x_p$ from the same category, our goal is to robustly predict their expression labels despite real-world noise and identity variations. To maintain a unified optimization pipeline, LaCoVL-FER integrates our primary methodological designs with an established contrastive learning stream. 

Specifically, for each input branch, the proposed LGAE introduces geometric priors by deploying our newly designed Bi-branch Gated Cross Attention (BGCA) mechanism to achieve adaptive fusion of landmark-based geometric and visual appearance features. This encoding process concurrently yields expression-relevant features at two distinct granularity levels for both the anchor and positive branches: local expression tokens ($X_a, X_p$) and global context features ($Z_a^{cls}, Z_p^{cls}$). Subsequently, these representations are processed through the two parallel streams. In the visual stream, the local tokens $X_a$ and $X_p$ are passed into the baseline CSL module~\cite{ref12} to capture fine-grained cross-sample correspondences, thereby suppressing category-irrelevant identity variations. Concurrently, in the multi-modal stream, the global context features $Z_a^{cls}$ and $Z_p^{cls}$ are fed into our core VLES to yield expression-specific visual representations. Based on these representations, an Expression-Conditioned Prompting (ECP) mechanism dynamically adapts the textual features of fixed class-level prompts into instance-aware textual representations. Notably, the cross-sample CSL branch is activated exclusively during training to provide auxiliary fine-grained constraints. During inference, LaCoVL-FER requires only a single test image as input, processing its individual features through the LGAE and VLES while completely bypassing the paired CSL computation to ensure optimal deployment efficiency.

\subsection{Landmark-Guided Adaptive Encoder}
In complex real-world scenarios, factors such as occlusion, pose variation, and illumination changes often cause attention mechanisms relying solely on appearance features to suffer from attention redundancy. To alleviate this issue, we propose a Landmark-Guided Adaptive Encoder (LGAE), which introduces landmark-based geometric priors to guide the model to focus on key facial regions, thereby learning more robust expression-relevant features.

Given an input image $I_m$, where $m\in\{a,p\}$ denotes the anchor and positive samples, respectively, we adopt a two-stream backbone to extract multi-scale facial appearance and geometric features. Specifically, the appearance stream employs a pretrained IR50~\cite{ref55} network, while the geometry stream adopts a pretrained MobileFaceNet~\cite{ref45}:
\begin{equation}
\begin{aligned}
F^{(i)}_{m,\mathrm{ir}}&=\mathrm{Backbone}_{\mathrm{IR}}(I_m),\\
F^{(i)}_{m,\mathrm{face}}&=\mathrm{Backbone}_{\mathrm{Face}}(I_m),
\end{aligned}
\label{eq_1}
\end{equation}  
where $i\in\{1,2,3\}$ denotes different spatial scales. Instead of directly fusing the two types of features in a unified space, we model the interactions between appearance and geometric representations in the multi-scale feature space, which helps preserve the intrinsic properties of appearance textures and geometric structures.

\begin{figure*}[!t] 
\centering
\includegraphics[width=\textwidth,trim=20 20 22 20,clip]{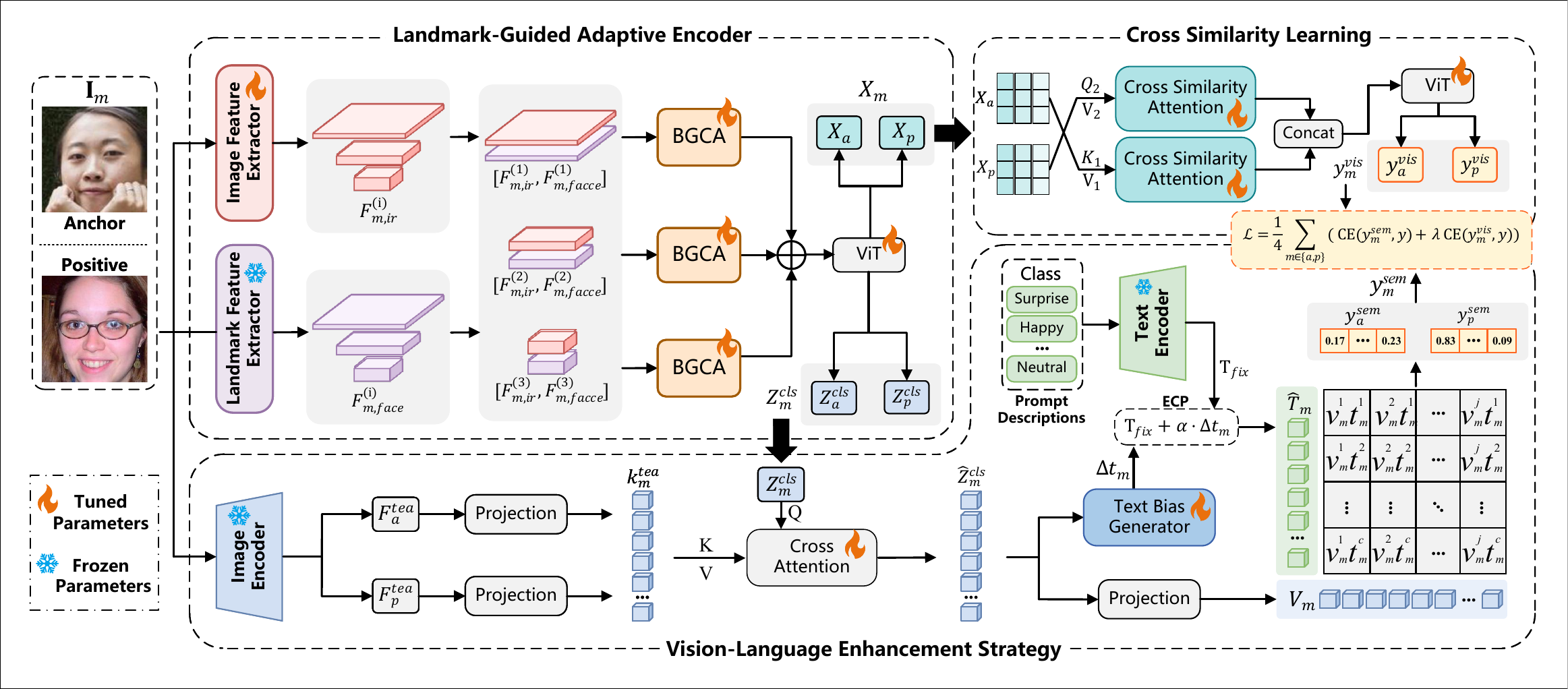}
\caption{Overview of the proposed LaCoVL-FER. To address attention redundancy and instability, it integrates two primary sample-adaptive refinement components: (1) the Landmark-Guided Adaptive Encoder (LGAE) introduces geometric priors via Bi-branch Gated Cross Attention (BGCA) for regional calibration, and (2) the Vision-Language Enhancement Strategy (VLES) employs expression-relevant feature refinement and an Expression-Conditioned Prompting (ECP) mechanism to align expression-specific visual and instance-aware textual representations as semantic priors, cooperatively optimized via the Cross Similarity Learning pathway during the training phase.}
\label{fig_2}
\end{figure*}

\begin{figure}[!t]
\centering
\includegraphics[width=3.5in,trim=18 18 18 20,clip]{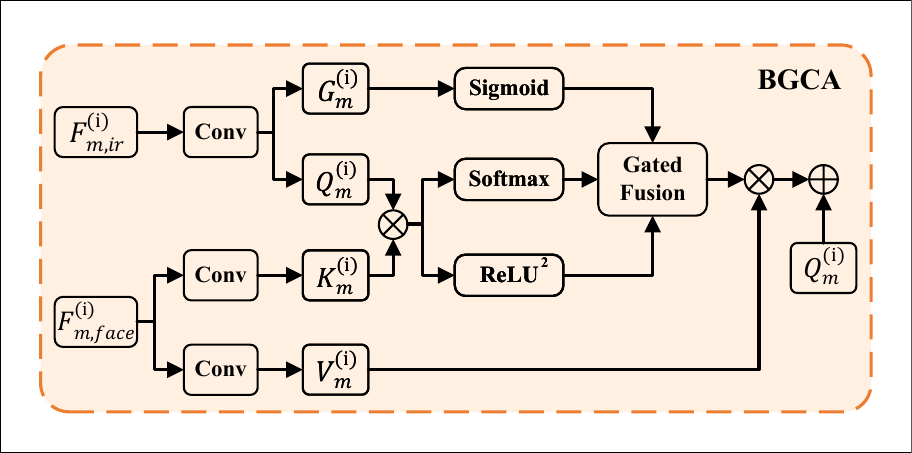}
\caption{The architecture of the BGCA mechanism at the $i$-th scale, which models the interaction between appearance features and landmark-based geometric features to produce the expression-relevant features.}
\label{fig_3}
\end{figure}

To effectively integrate geometric priors with visual appearance features, we design a Bi-branch Gated Cross Attention (BGCA) mechanism, as illustrated in Fig.~\ref{fig_3}. At each scale, BGCA treats the appearance feature $F^{(i)}_{m,\mathrm{ir}}$ as the primary visual representation, while the geometric feature $F^{(i)}_{m,\mathrm{face}}$ provides structural guidance. To enable effective interaction between the two modalities, the appearance feature is projected into query vectors and gating scores, while the geometric feature is projected into key and value vectors:
\begin{equation}
\begin{aligned}
Q^{(i)}_m,\,G^{(i)}_m &= \psi_q(F^{(i)}_{m,\mathrm{ir}};\Theta_q),\\
K^{(i)}_m &= \psi_k(F^{(i)}_{m,\mathrm{face}};\Theta_k),\\
V^{(i)}_m &= \psi_v(F^{(i)}_{m,\mathrm{face}};\Theta_v).
\end{aligned}
\label{eq_2}
\end{equation}

In the projected space, we compute the scaled dot-product similarity matrix as $S^{(i)}_{m} = {Q^{(i)}_{m}(K^{(i)}_{m})^T}/{\sqrt{d_h}}$. To simultaneously capture global dependencies and salient regional responses, BGCA decomposes the attention mechanism into two complementary branches: Dense Cross Attention (DCA) and Sparse Cross Attention (SCA). The DCA branch adopts standard Softmax normalization to capture global dependencies among features:
\begin{equation}
A^{(i)}_{m,\mathrm{dense}} =
\mathrm{Softmax}(S^{(i)}_{m}),
\label{eq_3}
\end{equation}

However, not all query-key pairs are equally relevant for expression discrimination. Directly aggregating all similarities may introduce redundant interactions and weaken the focus on critical regions. To address this issue, the SCA branch emphasizes high-response regions via a normalized $\mathrm{ReLU}(\cdot)^2$ activation, thereby selectively preserving more discriminative correlations:
\begin{equation}
A^{(i)}_{m,\mathrm{sparse}} =
\frac{\mathrm{ReLU}(S^{(i)}_{m})^2}
{\sum_{j=1}^{N}\mathrm{ReLU}(S^{(i)}_{m})^2 + \epsilon},
\label{eq_4}
\end{equation}
Compared with a standard ReLU activation, the squared operation further enlarges the contrast between high-response and low-response elements, thereby enhancing the discriminative capability of the attention distribution.

The deployment of parallel dense and sparse attention pathways exploits their complementary mathematical behaviors under wild facial variations. Specifically, the dense branch (DCA) preserves global holistic contexts for macro-level expressions, whereas the sparse branch (SCA) suppresses long-tail clutter to lock onto localized muscular structures. Crucially, instead of executing a static, hard-coded balancing ratio, our BGCA introduces an appearance-driven gating network to dynamically modulate their confluence. This adaptive mechanism directly mitigates the bottleneck of attention instability. By evaluating the environmental clarity and identity context of each input face, the gating network extracts an instance-specific score $G_m^{(i)}$, which is mapped to a dynamic weight $g_m^{(i)}$ via a Sigmoid activation $\sigma(\cdot)$. Based on this weight, a Gated Fusion operation is performed over the two branches to obtain the mixed attention map:
\begin{equation}
A_{m,\mathrm{mixed}}^{(i)} =
g_m^{(i)} \odot A_{m,\mathrm{dense}}^{(i)}
+
(1-g_m^{(i)}) \odot A_{m,\mathrm{sparse}}^{(i)}.
\label{eq_5}
\end{equation}

The mixed attention map is then used to aggregate the value vectors $V_m^{(i)}$ and projected back to the appearance feature space via $\psi_{\mathrm{proj}}(\cdot)$. To ensure training stability and alleviate gradient vanishing in deep networks, we further introduce residual connections and batch normalization, yielding the expression-relevant features:
\begin{equation}
\tilde{F}^{(i)}_{m,\mathrm{ir}} =
\mathrm{BN}\Big(
F^{(i)}_{m,\mathrm{ir}} +
\psi_{\mathrm{proj}}\!\left(
A^{(i)}_{m,\mathrm{mixed}} V^{(i)}_{m};\Theta_p
\right)
\Big).
\label{eq_6}
\end{equation}

After that, the expression-relevant features are projected into a unified token space and concatenated to form the fused multi-scale feature sequence, which is then fed into a Transformer encoder to generate the global class token $Z_m^{cls}$ and the local feature sequence $X_m$. To formulate the full optimization pipeline, the extracted local feature sequence $X_m$ is concurrently routed through the predefined Cross Similarity Learning pathway to yield the auxiliary vision-only prediction logits $y_m^{\mathrm{vis}}$.

\subsection{Vision-Language Enhancement Strategy}
To further exploit semantic priors from vision-language models, we introduce a Vision-Language Enhancement Strategy (VLES). Different from previous methods that rely only on fixed class-level prompts or limited learnable templates, the proposed VLES leverages the expression-relevant features extracted by the LGAE to refine the CLIP-derived generalizable visual features, generating expression-specific visual representations. Based on these representations, an Expression-Conditioned Prompting (ECP) mechanism is further employed to adapt the CLIP-derived textual features of fixed class-level prompts, thereby producing instance-aware textual representations. These visual-textual representations are aligned as semantic priors to enhance the robustness and generalization of FER.

\textit{1) Expression-Relevant Feature Refinement:}
Given an input image $I_m$, the LGAE produces the corresponding global class token $Z_m^{cls}$ from expression-relevant features, where $m\in\{a,p\}$. The generalizable visual feature $F_m^{tea}$ is extracted by the frozen pretrained CLIP image encoder $\mathcal{E}_{tea}(I_m)$ and projected into the current visual feature space through a linear mapping $\psi_{tea}(\cdot)$ to construct the corresponding key and value representation $k_m^{tea}$. Based on this, the global token $Z_m^{cls}$ is taken as the query, and cross-attention is applied to yield expression-specific visual representations:
\begin{equation}
\tilde{Z}_m^{cls} =
\mathrm{Attn}_{\mathrm{cross}}
\big(Q=Z_m^{cls},\;K=k_m^{tea},\;V=k_m^{tea}\big).
\label{eq_13}
\end{equation}

\textit{2) Expression-Conditioned Prompting:}
To alleviate the limitation of fixed prompts in describing complex expression variations, we introduce an Expression-Conditioned Prompting (ECP) mechanism. Let $\mathcal{P}=\{p_c\}_{c=1}^{C}$ denote the fixed class-level prompts for the $C$ expression categories. These prompts are encoded by the frozen pretrained CLIP text encoder $\mathcal{E}_{txt}(\cdot)$ to obtain the class-level textual representations:
\begin{equation}
T_{fix}=\mathrm{Norm}\big(\mathcal{E}_{txt}(\mathcal{P})\big).
\label{eq_14}
\end{equation}

Based on the expression-specific visual representations $\tilde{Z}_m^{cls}$, ECP introduces a Text Bias Generator to map $\tilde{Z}_m^{cls}$ into the textual semantic space and predict a textual semantic bias:
\begin{equation}
\Delta t_m=\tanh(W_{\delta}\tilde{Z}_m^{cls}+b_{\delta}),
\label{eq_15}
\end{equation}
where $W_{\delta}$ and $b_{\delta}$ denote the learnable parameters of the Text Bias Generator. This semantic bias is shared globally and used to consistently modulate the fixed class-level textual representations.

The final instance-aware textual representations are formulated as:
\begin{equation}
\hat{T}_m=\mathrm{Norm}(T_{fix}+\alpha\cdot\Delta t_m),
\label{eq_16}
\end{equation}
where $\alpha$ is a scaling factor that balances the contributions of fixed textual semantic priors and instance-specific textual semantic bias. In our experiments, $\alpha$ is empirically set to 0.1.

\textit{3) Visual--Textual Alignment:}
The expression-specific visual representations $\tilde{Z}_m^{cls}$ are projected into the textual semantic space to obtain the corresponding visual embedding:
\begin{equation}
V_m=\mathrm{Norm}\big(\psi_{img}(\tilde{Z}_m^{cls})\big),
\label{eq_17}
\end{equation}
where $\psi_{img}(\cdot)$ denotes the visual projection function that maps visual representations into the semantic space aligned with textual embeddings.

Based on the visual embedding and the instance-aware textual representations generated by ECP, semantic prediction logits are computed via cosine similarity:
\begin{equation}
y_m^{sem}=\frac{1}{\tau}\langle V_m,\hat{T}_m\rangle,
\label{eq_18}
\end{equation}
where $\tau$ is a temperature parameter used to control the smoothness of the similarity distribution.

\begin{algorithm}[t]
\caption{Training Pipeline of LaCoVL-FER}
\label{algorithm_1}
\SetKwInOut{Input}{Input}
\SetKwInOut{Require}{Require}
\SetKwInOut{Output}{Output}
\Input{Training set $\mathcal{D}_{train}=\{(I_a, I_p, y)\}$, fixed class-level prompts $\mathcal{P} = \{p_c\}_{c=1}^C$.}
\Require{Frozen pretrained CLIP image encoder $\mathcal{E}_{tea}$ and text encoder $\mathcal{E}_{txt}$, pretrained weights for IR50 and MobileFaceNet, and initialized model parameters $\Theta$.}
\Output{Optimized model parameters $\Theta^{*}$}
\For{$iter \leftarrow 1$ \KwTo $T_{\max}$}{
    Sample a mini-batch of anchor-positive pairs $(I_a, I_p, y)$ from $\mathcal{D}_{train}$\;
    \textbf{\# Landmark-Guided Adaptive Encoder}\;
    Extract multi-scale appearance and geometric features $\{F^{(i)}_{m,\mathrm{ir}}, F^{(i)}_{m,\mathrm{face}}\}$ for $m \in \{a,p\}$ according to Eq.~(\ref{eq_1})\;
    Compute expression-relevant features $\tilde{F}^{(i)}_{m,\mathrm{ir}}$ according to Eqs.~(\ref{eq_2})--(\ref{eq_6})\;
    Feed the fused multi-scale features into the Transformer encoder to obtain local token sequences $X_a, X_p$ and class tokens $Z_a^{cls}, Z_p^{cls}$\;
   \textbf{\# Cross Similarity Learning}\;
    Feed the local token sequence $X_m$ through the predefined Cross Similarity Learning pathway to yield the auxiliary vision-only prediction logits $y_a^{\mathrm{vis}}$ and $y_p^{\mathrm{vis}}$\;
    \textbf{\# Vision-Language Enhancement Strategy}\;
    Extract generalizable visual features via $\mathcal{E}_{tea}$ and refine them to obtain expression-specific visual representations $\tilde{Z}_a^{cls}$ and $\tilde{Z}_p^{cls}$ according to Eq.~(\ref{eq_13})\;
    Compute class-level textual representations $T_{fix}$ according to Eq.~(\ref{eq_14})\;
    Apply the ECP mechanism to generate textual semantic biases $\Delta t_m$ and instance-aware textual representations $\hat{T}_m$ according to Eqs.~(\ref{eq_15})--(\ref{eq_16})\;
    Project $\tilde{Z}_a^{cls}$ and $\tilde{Z}_p^{cls}$ into the textual semantic space and compute semantic logits $y_a^{\mathrm{sem}}$ and $y_p^{\mathrm{sem}}$ according to Eqs.~(\ref{eq_17})--(\ref{eq_18})\;
    \textbf{\# Optimization}\;
    Compute the total training loss $\mathcal{L}_{\mathrm{train}}$ according to Eq.~(\ref{eq_19})\;
    Update model parameters $\Theta \rightarrow \Theta^{*}$ by minimizing $\mathcal{L}_{\mathrm{train}}$ via backpropagation\;
}
\end{algorithm}

\begin{algorithm}[t]
\caption{Inference Pipeline of LaCoVL-FER}
\label{algorithm_2}
\SetKwInOut{Input}{Input}
\SetKwInOut{Require}{Require}
\SetKwInOut{Output}{Output}
\Input{Test set $\mathcal{D}_{test}=\{I_a\}$, fixed class-level prompts $\mathcal{P} = \{p_c\}_{c=1}^C$.}
\Require{Frozen pretrained CLIP image encoder $\mathcal{E}_{tea}$ and text encoder $\mathcal{E}_{txt}$, pretrained weights for IR50 and MobileFaceNet, and trained model parameters $\Theta^{*}$.}
\Output{Predicted expression label $\hat{y}$}
\ForEach{$I_a \in \mathcal{D}_{test}$}{
    \textbf{\# Landmark-Guided Adaptive Encoder}\;
    Extract multi-scale appearance and geometric features $\{F^{(i)}_{a,\mathrm{ir}}, F^{(i)}_{a,\mathrm{face}}\}$ according to Eq.~(\ref{eq_1})\;
    Compute expression-relevant features $\tilde{F}^{(i)}_{a,\mathrm{ir}}$ according to Eqs.~(\ref{eq_2})--(\ref{eq_6})\;
    Feed the fused multi-scale features into the Transformer encoder to obtain the class token $Z_a^{cls}$\;
    \textbf{\# Vision-Language Enhancement Strategy}\;
    Extract generalizable visual features via $\mathcal{E}_{tea}$ and refine them to obtain expression-specific visual representations $\tilde{Z}_a^{cls}$ according to Eq.~(\ref{eq_13})\;
    Compute class-level textual representations $T_{fix}$ according to Eq.~(\ref{eq_14})\;
    Apply the ECP mechanism to generate textual semantic bias $\Delta t_a$ and instance-aware textual representations $\hat{T}_a$ according to Eqs.~(\ref{eq_15})--(\ref{eq_16})\;
    Project $\tilde{Z}_a^{cls}$ into the textual semantic space and compute semantic logits $y_a^{\mathrm{sem}}$ according to Eqs.~(\ref{eq_17})--(\ref{eq_18})\;
    \textbf{\# Predict}\;
    Predict the final label $\hat{y}$ according to Eq.~(\ref{eq_20})\;
}
\end{algorithm}

\subsection{Overall Learning Objectives}
During training, LaCoVL-FER is optimized with anchor-positive pairs from the same expression class. For each branch $m\in\{a,p\}$, the semantic logits $y_m^{\mathrm{sem}}$ are produced by the proposed vision-language pathway, while the auxiliary visual logits $y_m^{\mathrm{vis}}$ are obtained from the CSL branch. The overall training loss is formulated as:
\begin{equation}
\mathcal{L}_{\mathrm{train}}
=
\frac{1}{4}
\sum_{m\in\{a,p\}}
\left(
\mathrm{CE}\!\left(y_m^{\mathrm{sem}}, y\right)
+
\lambda\,\mathrm{CE}\!\left(y_m^{\mathrm{vis}}, y\right)
\right),
\label{eq_19}
\end{equation}
where $\lambda$ is a trade-off parameter, empirically set to 1, and $y$ denotes the ground-truth label shared by both anchor and positive samples.

During inference, only the anchor pathway is retained, as cross-sample interaction is unnecessary for single-image prediction. Accordingly, the model achieves efficient inference without compromising the learned semantic representation capability. The final prediction is obtained from the semantic branch:
\begin{equation}
\hat{y}=\arg\max\left(y_a^{\mathrm{sem}}\right).
\label{eq_20}
\end{equation}

The detailed procedures of training and inference are described in Algorithm~\ref{algorithm_1} and Algorithm~\ref{algorithm_2}, respectively.

\section{Experiments}
\subsection{Datasets and Evaluation Metrics}
We conduct experiments on several widely used FER benchmarks, including RAF-DB~\cite{ref49}, FERPlus~\cite{ref50}, AffectNet~\cite{ref51}, as well as their corresponding occlusion and pose-variant subsets~\cite{ref8}, namely Occlusion-RAF-DB, Occlusion-FERPlus, Occlusion-AffectNet, Pose-RAF-DB, Pose-FERPlus, and Pose-AffectNet.

\textbf{RAF-DB}~\cite{ref49} is a real-world facial expression database containing 29,672 facial images annotated with basic or compound expressions. Following the standard evaluation protocol, we select 15,339 images annotated with seven basic expressions (surprise, fear, disgust, happiness, sadness, anger, and neutral). Among them, 12,271 images are used for training and 3,068 images for testing.

\textbf{FERPlus}~\cite{ref50} is an extension of FER2013~\cite{ref52}. Each image is re-annotated by 10 crowd-sourced annotators with eight expression labels (seven basic expressions plus contempt). The dataset consists of 28,709 training images and 3,589 test images. In our experiments, we follow the official FER2016 protocol with majority voting, where a single dominant label is retained for each image. Samples labeled as ``unknown'' or ``not a face (NF)'' are excluded. The recognition accuracy is reported on the test set.

\textbf{AffectNet}~\cite{ref51} is a large-scale in-the-wild FER dataset containing approximately 440,000 manually annotated facial images. We adopt both AffectNet-7 and AffectNet-8 settings. AffectNet-7 includes 283,901 training images and 3,500 validation images, while AffectNet-8 contains 287,651 training images and 4,000 validation images. To mitigate the training set's long-tailed distribution, balanced sampling~\cite{ref53} is employed. Conversely, the validation set is inherently class-balanced and thus used directly to ensure a fair evaluation.

\textbf{Occlusion-RAF-DB, Occlusion-FERPlus, and Occlusion-AffectNet}~\cite{ref8} are occlusion-specific facial expression subsets constructed from the RAF-DB test set, the FERPlus test set, and the AffectNet validation set, respectively. These subsets contain facial images with visible occlusions. Specifically, Occlusion-RAF-DB, Occlusion-FERPlus, and Occlusion-AffectNet include 735, 605, and 682 occluded facial images, respectively.

\textbf{Pose-RAF-DB, Pose-FERPlus, and Pose-AffectNet}~\cite{ref8} are pose-variant facial expression subsets derived from the RAF-DB test set, the FERPlus test set, and the AffectNet validation set. According to the pitch or yaw angle of the face, the samples are divided into two pose ranges: $\geq 30^\circ$ and $\geq 45^\circ$. Pose-RAF-DB contains 1,248 and 558 samples under the two pose ranges, respectively. Pose-FERPlus contains 1,171 and 634 samples. Pose-AffectNet includes 1,949 images with pose $\geq 30^\circ$ and 985 images with pose $\geq 45^\circ$.

\subsection{Implementation Details}
We adopt DCS~\cite{ref12} as the baseline and build our model upon it. All face images are resized to $224\times224$ before being fed into the network. During training, data augmentation techniques including random cropping, random erasing, random horizontal flipping, and color jittering are applied to improve generalization. The image backbone is IR50~\cite{ref55}, pretrained on the MS-Celeb-1M dataset~\cite{ref54}, and is further fine-tuned during training. For geometric feature extraction, a pretrained MobileFaceNet~\cite{ref45} with frozen weights is employed as the facial landmark detector to provide stable geometric priors. The VLES module is built upon CLIP (ViT-B/16)~\cite{ref24}, where the temperature parameter $\tau$ is set to 0.07. The model is optimized using the Adam optimizer~\cite{ref56}. The number of training epochs is set to 200 for RAF-DB, 150 for FERPlus, and 100 for AffectNet. The batch size is 24, the initial learning rate is 4e-6, and the weight decay is 1e-4. For the ECP module, the scaling factor $\alpha$ is set to 0.1. All experiments are conducted on an NVIDIA RTX 5090 GPU. Each experiment is repeated five times, then the maximum accuracy is reported.

\setlength{\heavyrulewidth}{1.2pt}
\setlength{\lightrulewidth}{0.6pt}
\begin{table*}[t]
\centering
\small
\setlength{\tabcolsep}{6pt}
\renewcommand{\arraystretch}{1.15}
\begin{threeparttable}
\caption{Comparison (\%) With State-of-the-Art Methods on RAF-DB, FERPlus, AffectNet-7 and AffectNet-8 Datasets.}
\label{tab_1}
\begin{tabular}{c|c|c|c|c|c|c|c}
\toprule
 & Methods & Year & \makecell{Vision\\Backbone} & RAF-DB & FERPlus & AffectNet-7 & AffectNet-8\\
\midrule
\multirow{10}{*}{\makecell{Unimodal:\\CNN-Based}}
& RAN~\cite{ref8}          & TIP20   & ResNet-18   & 86.90 & 89.16 & 59.50 & 52.97 \\
& SCN~\cite{ref38}         & CVPR20  & ResNet-18   & 87.03 & 88.01 & --    & 60.23 \\
& DMUE~\cite{ref59}        & CVPR21  & ResNet-18   & 88.76 & 88.64 & 62.84 & --    \\
& EfficientFace~\cite{ref39}& AAAI21 & ResNet-50   & 88.36 & --    & 63.70 & 59.89 \\
& EAC~\cite{ref60}         & ECCV22  & ResNet-18   & 89.99 & 89.64 & 65.32 & --    \\
& SOFT~\cite{ref61}        & ECCV22  & ResNet-18   & 90.42 & 88.60 & 66.13 & 62.69 \\
& FG-AGR~\cite{ref41}      & TCSVT23 & ResNet-18   & 90.81 & 91.09 & 64.91 & 60.69 \\
& MRAN~\cite{ref40}        & TCSVT23 & ResNet-18   & 90.03 & 89.59 & 66.31 & 62.48 \\
& NHG~\cite{ref67}         & PR26    & ResNet-18   & 90.09 & 88.94 & 65.14 & --    \\
& AUNet~\cite{ref72}       & KBS26   & ResNet-18   & 91.75 & 90.52 & 67.51 & --    \\
\midrule
\multirow{10}{*}{\makecell{Unimodal:\\CNN\&ViT-Based}}
& TransFER~\cite{ref11}    & ICCV21  & IR50+ViT-S  & 90.91 & 90.83 & 66.23 & --    \\
& VTFF~\cite{ref10}        & TAC21   & ResNet18+ViT-B & 88.14 & 88.81 & 64.80 & 61.85 \\
& APViT~\cite{ref58}       & TAC22   & IR50+ViT-S  & 91.98 & 90.86 & 66.91 & --    \\
& PIDVIT~\cite{ref62}      & TAC22   & ViT         & 90.71 & --    & 65.80 & 62.52 \\
& POSTER~\cite{ref15}      & ICCV23  & IR50+ViT    & 92.05 & 91.62 & 67.31 & 63.34 \\
& ExpLLM~\cite{ref63}      & TMM25   & ViT-L/14      & 91.03 & --    & 65.93 & 62.86 \\
& POSTER++~\cite{ref16}    & PR25    & IR50+ViT    & 92.21 & --    & 67.49 & 63.77 \\
& DCS~\cite{ref12}         & AAAI25  & IR50+ViT    & 92.57 & 91.41 & 67.66 & 64.40 \\
& QCS~\cite{ref12}         & AAAI25  & IR50+ViT    & 92.50 & 91.41 & 67.94 & 64.30 \\
& AMGSN~\cite{ref68}       & PR26    & ViT-B/16    & 89.34 & --    & 62.83 & --    \\
\midrule
\multirow{7}{*}{\makecell{Multimodal:\\CLIP-Based}}
& CLIPER~\cite{ref24}      & ICME24  & CLIP-ViT-B/16 & 91.61 & --    & 66.29 & 61.98 \\
& CEPrompt~\cite{ref25}    & TCSVT24 & CLIP-ViT-B/16 & 92.43 & --    & 67.29 & 62.74 \\
& VLCA~\cite{ref30}        & ESWA25  & CLIP-ViT-B/32 & 92.12 & --    & 65.84 & 61.96 \\
& TPRD~\cite{ref31}        & TAC25   & CLIP-ViT-B/16 & 92.89 & 91.01 & 68.00 & 62.85 \\
& ICoCO~\cite{ref69}       & TAC26   & CLIP-ViT-B/16 & 92.57 & 90.25 & 67.66 & 62.98 \\
& MMPL-FER~\cite{ref32}    & IF26    & CLIP-ViT-B/32 & 93.28 & 91.49 & 67.69 & --    \\
& \textbf{LaCoVL-FER (Ours)}   & 2026      & CLIP-ViT-B/16 & \textbf{93.61} & \textbf{91.79} & \textbf{68.14} & \textbf{64.75} \\
\bottomrule
\end{tabular}
\end{threeparttable}
\end{table*}

\begin{figure*}[!t]
\centering
\includegraphics[width=7in,trim=9 8 8 8,clip]{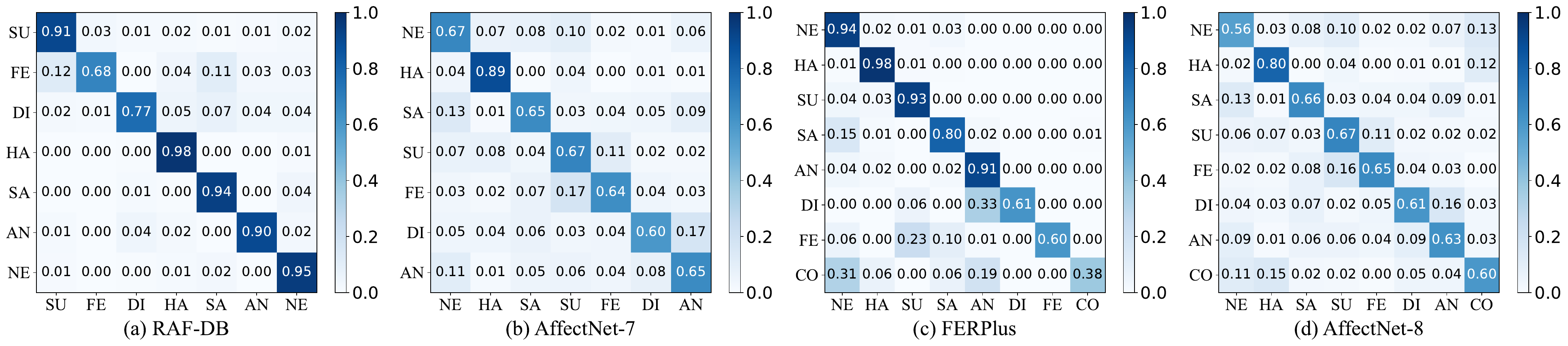}
\caption{Confusion matrices of our LaCoVL-FER model, where SU: surprise, FE: fear, DI: disgust, HA: happy, SA: sad, AN: anger, NE: neutral, CO: contempt.}
\label{fig_5}
\end{figure*}

\subsection{Comparison With State-of-the-Arts}
Table~\ref{tab_1} compares LaCoVL-FER with state-of-the-art FER methods on RAF-DB, FERPlus, AffectNet-7, and AffectNet-8. Overall, LaCoVL-FER achieves the best performance across all benchmark settings, with accuracies of 93.61\%, 91.79\%, 68.14\%, and 64.75\%, respectively. Compared with CNN-based methods, which mainly rely on local appearance patterns, LaCoVL-FER benefits from adaptive landmark-guided regional calibration. Compared with CNN\&ViT-based methods, which improve long-range visual modeling but still operate mainly in the visual domain, LaCoVL-FER further introduces expression-conditioned semantic priors from a vision-language model. Compared with recent CLIP-based FER methods, our framework does not directly rely on fixed visual-textual representations. 
Instead, it refines CLIP-derived visual features into expression-specific representations and adapts textual prompts according to each input face, leading to more reliable visual-semantic alignment.

To further examine class-wise behavior, Fig.~\ref{fig_5} presents the confusion matrices on the three benchmarks. LaCoVL-FER produces clear diagonal dominance on RAF-DB, indicating stable predictions for most basic expressions. On FERPlus and AffectNet, the remaining errors mainly occur between visually similar categories, such as fear/surprise and disgust/anger, where the discriminative facial muscle movements are subtle and often localized. We also observe that the recognition rate of ``sad'' is relatively low across the three datasets. 
This is a common and explainable difficulty in in-the-wild FER: sad expressions are usually characterized by weak facial deformations and can easily overlap with neutral or negative expressions such as anger and fear, especially when the mouth and eyebrow cues are not sufficiently pronounced. Nevertheless, LaCoVL-FER maintains balanced performance across categories and reduces severe category collapse, suggesting that the adaptive use of landmark geometry and vision-language semantics improves both overall recognition accuracy and class-level discriminability.

\begin{figure}[!t]
\centering
\includegraphics[width=3.5in,trim=25 18 19 23,clip]{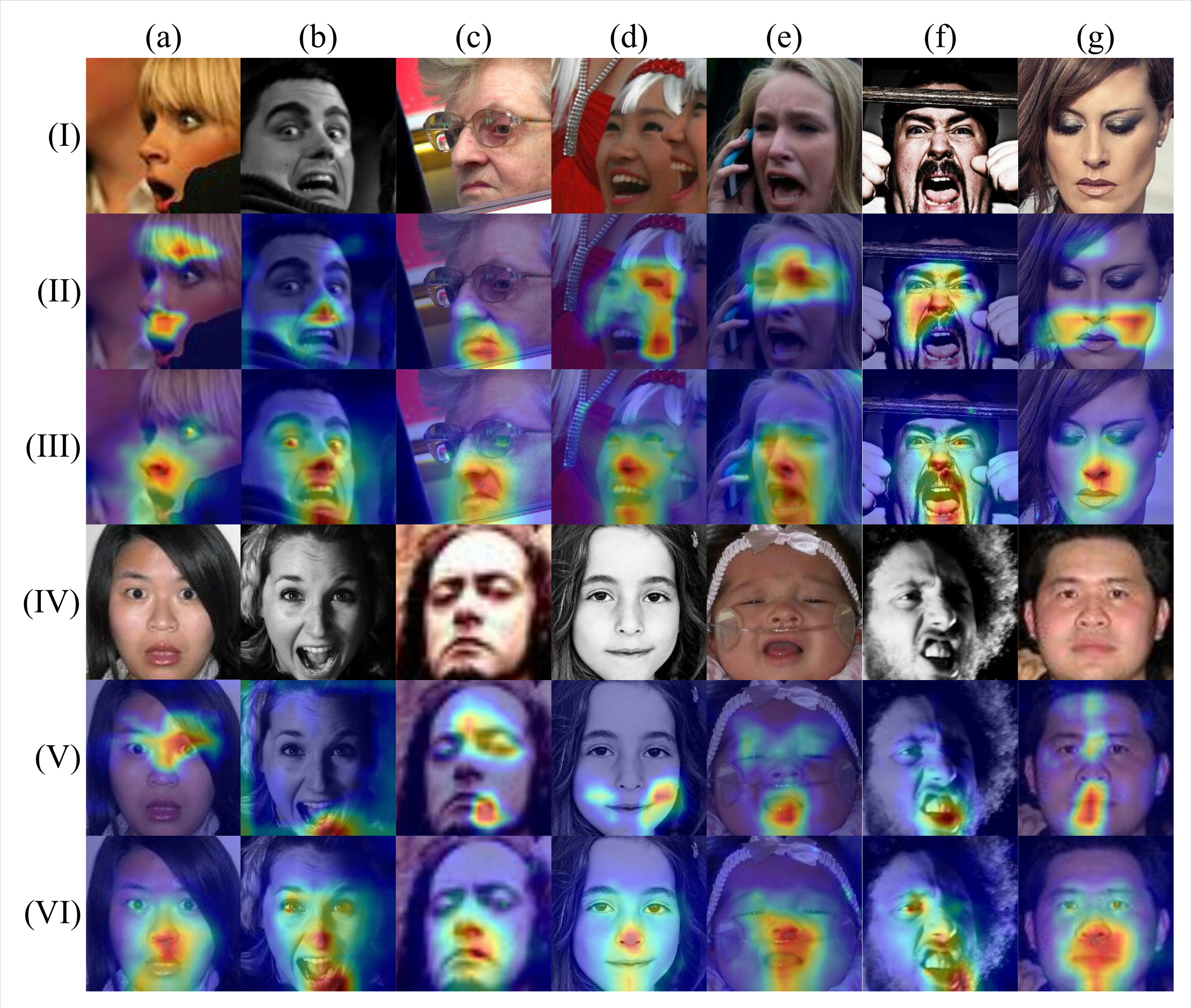}
\caption{Attention visualization for images from RAF-DB dataset. Column (a)--(g) denote “surprise”, “fear”, “disgust”, “happy”, “sad”, “anger”, “neutral”, respectively. Rows (I) and (IV) present original face images. Rows (II) and (V) illustrate the unstable and redundant attention maps of DCS, while Rows (III) and (VI) show the corresponding attention maps of LaCoVL-FER, which are more stable and focused.}
\label{fig_6}
\end{figure}

\subsection{Visualization Analysis}
\textit{1) Attention Visualization:} As illustrated in Fig.~\ref{fig_6}, we visualize the attention distributions on the RAF-DB test set. To demonstrate the consistency of the results, two sets of examples are provided, corresponding to Rows (I)--(III) and Rows (IV)--(VI), respectively. Rows (I) and (IV) display the original images. Rows (II) and (V) show the attention maps of the baseline DCS. Relying primarily on visual appearance cues, DCS suffers from attention redundancy and instability in complex scenarios. For instance, in column (d) of Row (II), its attention is erroneously diverted to the background face; meanwhile, a comparison between column (f) of Row (II) and column (f) of Row (V) clearly reveals that its responses to key facial regions are highly unstable. In contrast, Rows (III) and (VI) present the attention maps of LaCoVL-FER. By exploiting the BGCA mechanism for the adaptive fusion of landmark-based geometric and visual appearance features, LaCoVL-FER exhibits a much more stable and precisely focused attention pattern. It effectively suppresses background interference (as shown in column (d) of Row (III)) and consistently focuses on key facial regions (as evidenced by the comparison between column (f) of Row (III) and column (f) of Row (VI)). These visualizations intuitively demonstrate that BGCA successfully guides the model to produce robust expression-relevant features.

\begin{figure}[!t]
\centering
\includegraphics[width=3.5in,trim=91 23 83 18,clip]{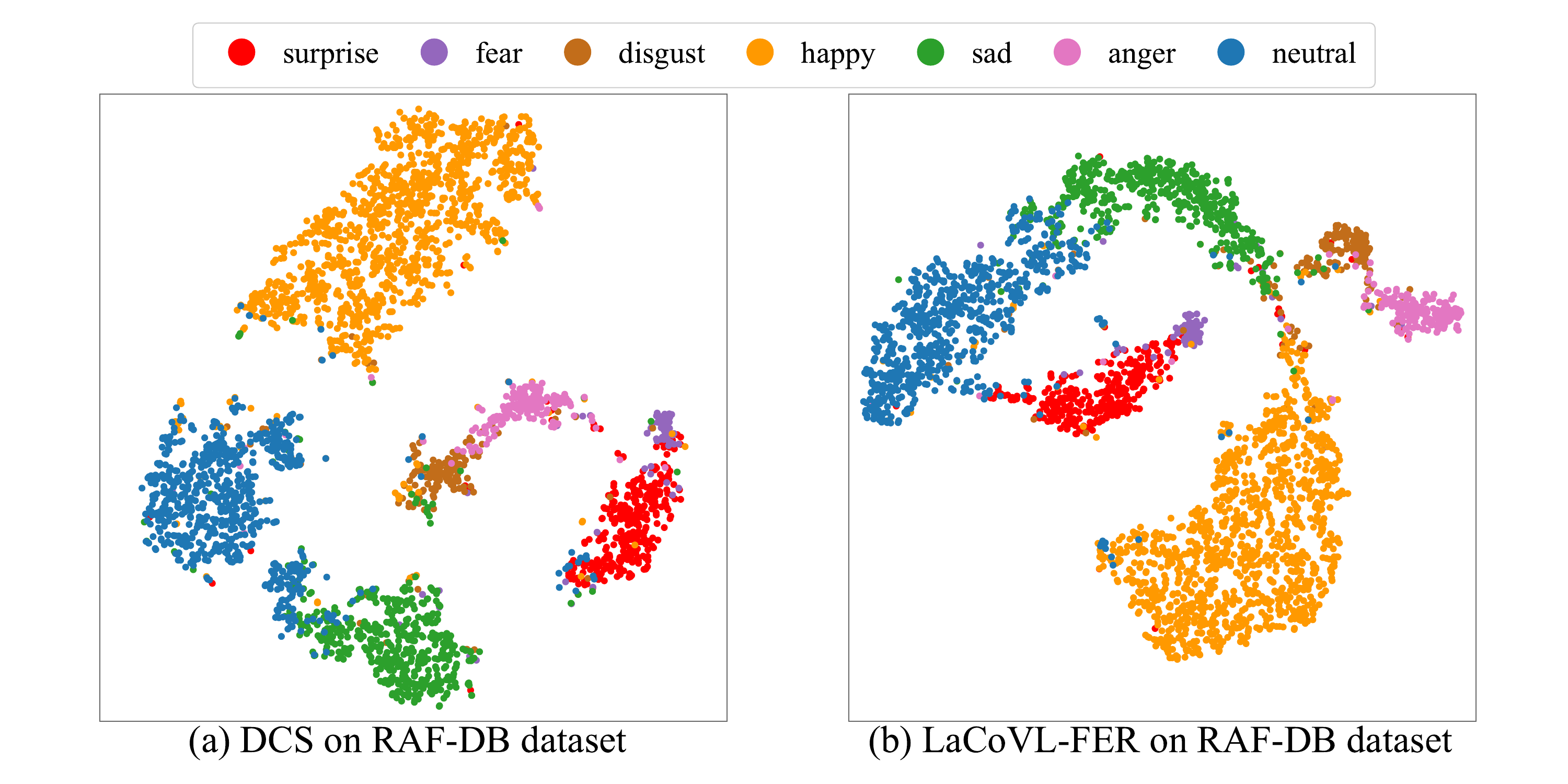}
\caption{Feature visualization using t-SNE~\cite{ref57} on RAF-DB dataset.}
\label{fig_7}
\end{figure}

\textit{2) Feature Visualization:}
As illustrated in Fig.~\ref{fig_7}, we further visualize the feature distributions on the RAF-DB test set using t-SNE~\cite{ref57}. As observed in Fig.~\ref{fig_7}(a), although the baseline DCS increases the overall inter-class distances, its feature space exhibits distinct fragmentation. Specifically, the ``neutral'' category is partitioned into multiple discontinuous sub-clusters, where some detached sub-clusters are located in close proximity to the ``sad'' category in the feature space, while maintaining a distinct gap from their own main cluster, reflecting substantial intra-class variance and an unstable feature structure. In contrast, the feature space of LaCoVL-FER in Fig.~\ref{fig_7}(b) demonstrates high cohesiveness. Benefiting from the effective constraints of both geometric and semantic priors, samples within each expression category shrink towards their respective centers, forming more compact clusters with clearer boundaries. Simultaneously, the confusion between visually similar categories in the boundary regions is significantly reduced, demonstrating that the proposed method learns more stable and discriminative feature representations.

\section{Conclusion}
In this paper, we present LaCoVL-FER, a novel landmark-guided contrastive learning framework with vision-language enhancement for robust in-the-wild facial expression recognition. Our work addresses the key limitation of existing methods, where visual, geometric, and semantic cues are often utilized in a static or weakly adaptive manner without sample-specific adaptation, inevitably introducing attention redundancy and representation instability. Specifically, the proposed Landmark-Guided Adaptive Encoder (LGAE) leverages the Bi-branch Gated Cross Attention (BGCA) mechanism to calibrate regional appearance features with facial landmark geometry, adaptively suppressing noisy responses and yielding expression-relevant visual representations. Complementarily, the Vision-Language Enhancement Strategy (VLES) refines generalizable CLIP visual features into expression-specific representations, while the Expression-Conditioned Prompting (ECP) mechanism dynamically adapts fixed class-level textual prompts into instance-aware semantic priors. Extensive experiments on RAF-DB, FERPlus, and AffectNet demonstrate that LaCoVL-FER outperforms state-of-the-art methods, achieving more discriminative and stable representations for real-world FER. Future work will extend the model to video-based FER and complex open-world applications.

\end{document}